# Prompt, Plan, Extract: Zero-Shot Agentic LLMs Workflows for Lung Pathology Extraction from Clinical Narratives

**Aman Pathak, MS[1], Cheng Peng, PhD [1], Mengxian Lyu, MS[1], Ziyi Chen, MS[1], Reema Solan, MS[1], Sankalp Talankar, MS[1], Yasir Khan, MS[1], Hiren Mehta, MD[2], Aokun Chen, PhD[3], Yi Guo, PhD[1], Yonghui Wu, PhD[1]**

**[1]Department of Health Outcomes and Biomedical Informatics, College of Medicine, University of Florida, Gainesville, FL, USA**
**[2]Division of Pulmonary, Critical Care and Sleep Medicine, Department of Medicine, College of Medicine, University of Florida, Gainesville, FL, USA**
**[3]College of Nursing, Florida State University, Tallahassee, FL, USA**

**Abstract**
*Information extraction from pathology reports is essential for cancer staging, tumor registry population. Yet key data remains embedded in narrative reports, making manual extraction labor-intensive and error-prone. Traditional supervised Natural Language Processing pipelines address this through fully supervised Named Entity Recognition and Relation Extraction, but require expensive manual annotation and suffer cascading failures when upstream entities are missed. In this study, we developed a zero-shot, agentic workflow, and evaluated five open-source generative Large Language Models (LLMs) to populate 13 College of American Pathologists synoptic fields from lung resection pathology reports. We compared them against a state-of-the-art supervised GatorTron NER-RE baseline using a novel, registry-aligned evaluation framework. The baseline achieved Micro-F1of 0.960, while the best zero-shot model (GPT-OSS-20B) achieved Micro-F1 of 0.893 (recall: 0.949), accurately extracting complex relations like Pathologic Stage without task-specific training. These results suggest that open-source, zero-shot agentic LLMs show great potential as a low-cost solution for extracting lung pathology information.*

**Introduction**
The Learning Health System (LHS) relies on the seamless conversion of clinical experience into actionable data to foster a continuous cycle of improvement in patient care and outcomes. Within oncology, pathology reports are an important source for diagnostic decisions, prognostic staging, and therapeutic guidance. However, much of the clinical detail remains locked in narrative text, such as pathology reports[1]. In lung cancer, for example, pathologic staging typically requires manual synthesis of narrative or semi-structured text regarding maximum tumor size, invasion scope, and the precise anatomical location of lymph node metastases. This manual process is not only labor-intensive but also highly prone to human error [2]. At scale, these manual workflows consume significant efforts of clinicians and coordinators. One recent evaluation of HER-to-EDC automation in oncology trials estimated that automating even half of the required fields could save up to $15,000 per patient in data-collection costs.[3] Furthermore, as reporting guidelines continuously evolve, the reliance on manual chart review to update or verify extractions causes substantial delays in clinical decision-making and fundamentally limits the scalability of real-world evidence generation.

Recent advances in clinical Natural Language Processing (NLP), especially large language models (LLMs) and generative artificial intelligence (AI), have revolutionized information extraction (IE) from clinical narratives.[4] The current best practice of IE is based on traditional extraction-driven solutions, e.g., Named Entity Recognition (NER). In traditional NER, IE is formulated as a two-stage pipeline, NER followed by Relation Extraction (RE).[5] Researchers typically fine-tune open-source LLMs, such as BERT[6], or clinical LLMs such as GatorTron[7,8], on manually annotated clinical corpora. In our previous work on extracting thyroid nodule characteristics from ultrasound reports, we demonstrated that this paradigm could achieve strong performance[9]. The GatorTron-based pipeline extracted 16 thyroid nodule characteristics from ultrasound reports with a strict F1 of 0.885, outperforming other rule-based and transformer baselines. Previous studies have also identified bottlenecks in the two-stage solutions for IE. First, it requires large volumes of manually annotated training data, a resource that is expensive to produce and domain-specific, limiting generalizability. Second, and more critically, the sequential design is prone to "cascade errors": when the NER module fails to detect an entity span, the RE module has no candidate to operate on.

Recently, generative AI, a subset of LLMs based on decoder-only LLMs, has enabled zero-shot clinical IE, which does not require manual annotation of domain-specific corpora for fine-tuning. While approaches utilizing generative LLMs for zero-shot NER have shown promise, the literature shows their zero-shot performance remains highly inconsistent. Accuracy varies substantially by concept type; for instance, some models extract standard medications with an F1 score of ~86% but drop to ~54% for complex treatments. [10] Zero-shot relation extraction has proven even more challenging, yielding mixed results on complex sentences containing nested relationships [11]. As a result, zero-shot LLMs are promising but often inconsistent without additional scaffolding. Empirical evidence suggests that prompt engineering, specifically heuristic templates and task decomposition, is critical to stabilizing LLM outputs for highly structured extraction tasks [12]. While recent studies have demonstrated that generative LLMs can outperform supervised methods in extracting findings from lung cancer radiology reports [13], the extraction of highly structured, relational data governed by clinical guidelines, such as the College of American Pathologists (CAP) synoptic protocols, remains under-evaluated. Furthermore, traditional NLP benchmarks evaluate models based on exact text span overlaps, whereas true clinical registries require normalized, registry-ready field-value pairs.

To address this gap, this study proposes a zero-shot, agentic LLM workflow (orchestrated via LangGraph) to automatically populate 13 canonical CAP fields from narrative lung pathology reports. We systematically compare five open-source generative LLMs against a state-of-the-art supervised traditional NER-RE solution, employing a novel entity-level evaluation framework and qualitative error analysis to demonstrate how agentic workflows can improve the extraction of lung pathology information.

## Methods

### Data Source

This study used pathology reports from the University of Florida (UF) Health Integrated Data Repository (IDR), a clinical data warehouse aggregating clinical and administrative data across UF Health systems. Reports were identified under an IRB-approved protocol, IRB202003159, targeting lung resection pathology reports from 2018-2022. All reports were processed in accordance with UF Health data governance policies and de-identified.

### Annotation

We recruited 3 annotators to manually identify both entity spans and typed relations related to lung pathology. The extraction schema was hierarchically modeled after the College of American Pathologists (CAP) synoptic reporting guidelines for lung carcinoma, encompassing core domains such as Specimen Details, Tumor Characteristics, Margin Status, Lymph Nodes, and Pathologic Staging. Initial annotation guidelines were iteratively refined under the clinical guidance and adjudication of a domain expert (Hiren Mehta) to ensure precise operationalization of clinical rules. The clinical schema targeted 13 core structured fields (Figure 1) as follows:

1. Histologic Type: The specific pathological classification of the tumor (e.g., Invasive adenocarcinoma)
2. Histologic Grade: The degree of differentiation (e.g., G2: Moderately differentiated).
3. Tumor Site: The specific anatomical location of the tumor within the lung (e.g., Right upper lobe).
4. Laterality: The side of the body (Right/Left) associated with the specimen.
5. Focality: The multiplicity of the tumor (e.g., Unifocal, Multiple nodules).
6. Tumor Size: The greatest dimension of the invasive tumor component, measured in centimeters or millimeters.
7. Visceral Pleural Invasion (VPI): Presence or absence of invasion into the visceral pleura (PL1/PL2).
8. Lymphovascular Invasion (LVI): Presence or absence of tumor cells within lymphatic or vascular channels.
9. Margin Status: The overall status (Uninvolved/Involved) of surgical margins (bronchial, vascular, parenchymal).
10. Margin Distance: The quantitative distance from the invasive carcinoma to the closest surgical margin.
11. Lymph Nodes Examined Count: The total integer counts of regional lymph nodes harvested and examined.
12. Lymph Nodes Involved Count: The total integer counts of lymph nodes found to contain metastatic carcinoma.
13. Pathologic Stage: The final pathologic stage group (e.g., pT2a N0) derived from the synthesis of T (tumor), N (node), and M (metastasis) descriptors.

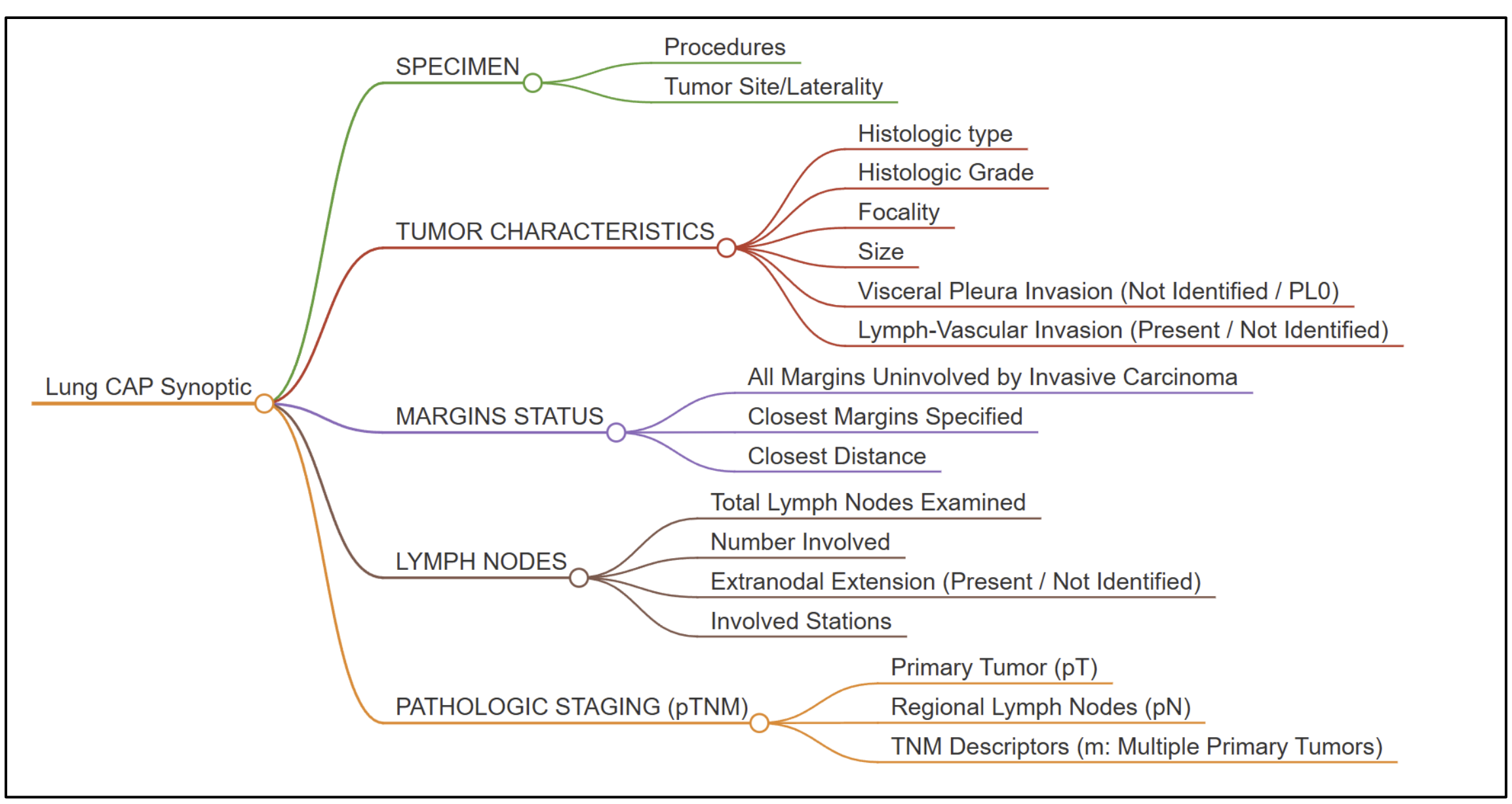

**Figure 1.** Mind-map of CAP-aligned clinical information extraction schema for lung resection pathology reports.

The extraction schema was hierarchically modeled after the College of American Pathologists (CAP) synoptic reporting guidelines for lung carcinoma, encompassing core domains such as Specimen Details, Tumor Characteristics, Tumor Extent and Invasion, Margin Status, Lymph Nodes, and Pathologic Staging. Initial annotation guidelines were iteratively refined under the clinical guidance and adjudication of a domain expert (Dr. Hiren Mehta). To ensure high corpus quality, three independent trained annotators underwent a rigorous, multi-phase calibration process. The team completed three iterative rounds of double-annotation and guideline refinement on overlapping subsets of pathology reports (ranging from 10 to 22 files per round). Performance was strictly monitored and by the final calibration round (conducted on 22 blinded, double-annotated reports), the annotators achieved an Entity F1-score of 0.843 (Cohen's Kappa = 0.810) and a Relation F1-score of 0.802, demonstrating strong, reliable agreement. All annotation discrepancies during the training phase were resolved through consensus meetings and expert adjudication before the annotators proceeded to label the remaining 286 pathology reports.

**Supervised Baseline: NER and Relation Extraction Pipeline**

To train the supervised baseline models, the annotated corpus was split into training, validation, and test sets using a 7:1:2 ratio. This partition allocated 58 reports to the test set. Upon quality review, 5 reports were excluded from the test set as they contained no relevant CAP schema annotations (empty gold standards), resulting in a final test set of 53 reports. This cohort was held out during the supervised training phase and served as the ground truth to evaluate the traditional NLP pipeline and the zero-shot generative LLMs.

To establish a rigorous, state-of-the-art traditional NER-RE-based NLP baseline, we implemented a two-stage sequential extraction pipeline consisting of Named Entity Recognition (NER) followed by Relation Extraction (RE). We used GatorTron345M [9], a large transformer-based clinical language model pretrained on billions of words from de-identified electronic health records, as the shared encoder backbone for both stages.

In the NER stage, we framed pathology extraction as a token-level sequence labeling task. Each report was tokenized to a maximum sequence length of 256, and the model was finetuned for 30 epochs using an AdamW-style optimizer. In the RE stage, we trained a separate model, also with a 256-token context window, to classify relations between pairs of annotated entities, thereby reconstructing header→value links (e.g., linking "1.5 cm" to "Tumor Size" or "uninvolved" to "Parenchymal Margin"). Both models were trained on an NVIDIA A100 40GB GPU. The approximate wall-clock training time was ~8 hours for NER and ~14 hours for RE.

For evaluation on the held-out test set (n=53), we executed a true end-to-end inference pipeline: raw text reports were first processed by the trained NER model to produce predicted entity spans, which were then passed to the trained RE model to infer relational pairs.

**Zero-Shot Agentic Workflow**

We designed a graph-structured, agentic workflow that uses generative LLMs to perform end-to-end clinical abstraction directly from raw text. This workflow was implemented using LangGraph and consists of four coordinated nodes: Mapper, Planner, Executor, and Compiler, that together map a free-text pathology report into a CAP-aligned JSON representation of the 13 target fields.

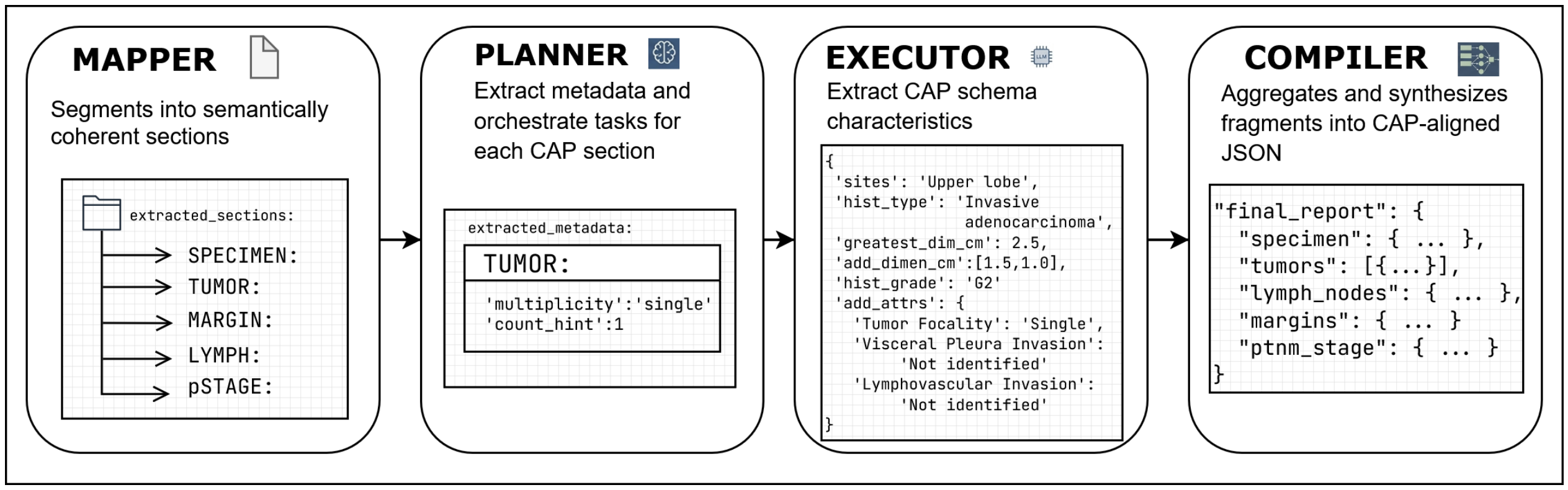

**Figure 2.** Architecture of the zero-shot LangGraph agentic workflow for CAP-aligned lung pathology extraction.

1. **Mapper:** The Mapper node receives the raw pathology report as input and segments it into semantically coherent sections such as SPECIMEN, TUMOR, MARGINS, LYMPH NODES, and PATHOLOGIC STAGING. This step anchors the extraction to the natural rhetorical structure of pathology reports and reduces the burden on later nodes by isolating relevant content for each clinical concept group.

2. **Planner**: The Planner uses the sections to extract high-level metadata and guide the executor. For each CAP schema section, it extracts different types of metadata and constructs a task queue of fine-grained extraction instructions, e.g., for the TUMOR section: it extracts tumor focality (single/multiple) and count hints, for the MARGINS section, it extracts involvement status for each margin and total margins mentioned. Since these tasks are highly independent, we enabled parallel planning of each CAP element. This step produces an ordered list of structured task descriptors that is passed to the Executor.

3. **Executor**: The Executor processes tasks in the task queue using specialized extraction functions. For each task, a prompt is constructed that includes (i) the relevant section text (from Mapper), (ii) the extracted metadata (from Planner), and (iii) a concise instruction to return a structured JSON response. E.g., For TUMOR section: the specialized extraction function will extract tumor characteristics (id, sites, histologic type, dimensions, grade). Similar to the planner, tasks are run concurrently, which reduces wall-clock latency while preserving isolation between tasks. The output is a list of per-task JSON fragments, which are passed to the Compiler.

4. **Compiler**: The Compiler node aggregates and synthesizes the fragments produced by the Executor into a single CAP-aligned JSON object. It merges overlapping observations, resolves simple conflicts (e.g., repeated mentions of the same size or margin status), and standardizes the outputs into a cohesive JSON containing the 13 canonical fields defined above.

This architecture transforms unstructured pathology reports into structured, actionable data, with each node having a focused responsibility, making the system easy to understand and maintain while handling the complexity and variability of medical language. For each backend, the Mapper, Planner, and Executor nodes used the same prompts and task definitions; only the model endpoint was changed. No in-context demonstrations or few-shot exemplars were provided at any stage; the models operated purely in a zero-shot regime.

**Prompt Design & Experimental Setup**

We manually engineered the zero-shot agentic workflow prompts to mirror the functional responsibilities of each LangGraph node, using role-specific system instructions and tightly scoped task prompts to minimize hallucination and maximize formatting reliability.

1. In the Mapper node, the model is instructed to behave as a "specialized medical report expert" and to segment each lung tumor pathology report into canonical clinical sections (SPECIMEN, TUMOR, MARGINS, LYMPH NODES, PATHOLOGIC STAGE), returning the full, unaltered section text without summarization or omission.
2. The Planner node operated as an "oncology clinical history analyst" which extracts high-level metadata used to guide downstream extraction, including tumor focality (single vs multiple), lymph node count hints, margin involvement planning cues, and neoadjuvant treatment/staging context (e.g., identifying whether a pTNM vs ypTNM prefix is expected). Critically, the Planner's outputs were passed to the Executor as structured context hints, reducing ambiguity in complex cases such as multi-focal tumors or post-neoadjuvant reports.
3. The Executor node performs section-specific structured extraction using dedicated prompts for specimen, tumor, margins, lymph nodes, and pTNM staging, with explicit type constraints (e.g., integer counts for lymph nodes, floats for tumor dimensions and margin distance) and instructions to return `null` when a value is absent. Across Executor prompts, we also implemented an "add_attrs" (short for: additional_attributes). This catch-all mechanism forces capture of any labeled key-value statements not covered by the core schema, reducing information loss during schema population.
4. The Compiler node aggregates the extracted JSON fragments into a single CAP-aligned, registry-ready JSON output.

Since the primary objective was to evaluate the baseline reasoning capabilities of different LLMs, the engineered prompts were held constant across all model backends. While the framework successfully operated across five distinct LLMs, this suggests a baseline level of prompt robustness. In this study, we did not formally evaluate the system's sensitivity to semantic variations in prompt phrasing. All evaluations were conducted in a strictly zero-shot setting (no in-context examples), with a temperature of 0.0 to encourage deterministic outputs. To improve robustness to transient formatting failures, each model call was configured with maximum 2 retries.

**Selection of Open-Source LLM Backends**

While the underlying LangGraph architecture is strictly model-agnostic, allowing for the seamless, drop-in replacement of any instruction-following model without altering the code, the overall extraction accuracy remains inherently sensitive to the chosen LLM's reasoning capabilities and parameter scale. To assess the performance of zero-shot open-source models, we compared gpt-oss-20b and gpt-oss-120b, two open-weight reasoning models released under an Apache-2.0 license and trained with large-scale distillation and reinforcement learning to support instruction-following and agentic use cases [14]. We also evaluated Gemma-3-27B-it, an instruction-tuned, open-weights model from the Gemma 3 family with long-context support (up to 128K tokens) intended for general-purpose text generation and reasoning [15,16]. In addition, we tested Llama-3.3-70B-Instruct, a 70B-parameter instruction-tuned model designed for assistant-style dialogue and strong general benchmark performance [17,18]. Finally, we included Granite-3.3-8B-Instruct, an 8B-parameter instruction-tuned model designed for long-context instruction following and lightweight deployment, representing a smaller-footprint alternative in the open-source model spectrum. [19]

**Entity-Level Evaluation Framework**

Traditional NER evaluation relies on text-level strict span matching, which rewards exact character offsets for every entity occurrence. This evaluation metric has been widely used for NER, reflecting performance in extracting text-based phrases, but it cannot accurately reflect performance in real-world lung pathology. Therefore, to reflect the real-world lung pathology, we used an evaluation framework derived from the CAP synoptic reporting guidelines.

First, the outputs from NER-RE-based NLP models were parsed into a unified logical representation consisting of (Field, Value) tuples, e.g. ("Tumor Site", "right upper lobe") or ("Histologic Grade", "moderately differentiated"). To account for generative variability, we implemented an aggressive string normalization layer that includes lowercasing, punctuation stripping, unit conversion (e.g. 70 mm to 7.0 cm), and synonym mapping for concepts (e.g., mapping "G2" to "moderately differentiated"). Second, to account for the redundant nature of clinical documentation, we transitioned from text-level to concept-level matching. In a standard pathology report, a single clinical fact (e.g., a tumor size of "1.5 cm") is often documented multiple times across the gross description, synoptic summary, and final diagnosis, resulting in multiple identical annotations. Because LLMs are designed to deduplicate these mentions automatically, standard NER evaluation would penalize LLMs with False Negatives for duplicated information. Therefore, the new evaluation is based on concept-level deduplication: identical normalized values for the same field within a report were treated as one to avoid penalizing LLMs. This new evaluation schema reflects the real-world lung pathology, as the same information only needs to be identified once.

Next, our exploratory data analysis of the manually annotated gold standard revealed high rates of annotator sparsity and ontological inconsistency. Human annotators frequently omitted explicitly negative findings (e.g., failing to tag Visceral Pleural Invasion when the text stated, "Not identified") or utilized generic relations rather than specific ones (e.g., linking a status to a generic "Margin" rather than the specific "Bronchial Margin"). To handle these variations, we introduced an "Implicit Null" forgiveness rule: if the gold standard lacked annotations for Lymph Node counts, Visceral Pleural Invasion, or Lymphovascular Invasion, AND a model explicitly extracted a valid negative status (e.g., "0", "Not identified", "Not Applicable"), the extraction was neutralized rather than penalized as a hallucinated False Positive. Additionally, to accommodate annotator inconsistency, highly granular margin predictions were collapsed into a unified "Margin Status" category for quantitative scoring.

**Results**

Across 53 held-out pathology reports, the supervised GatorTron NER-RE baseline achieved the highest overall performance (Micro-F1 0.960, Precision 0.944, Recall 0.977). All zero-shot LLM backends executed under the same LangGraph workflow achieved strong performance without any fine-tuning, ranging from Micro-F1 of 0.831-0.893 and recall of 0.924-0.957. The GPT-OSS-20B model emerged as the top-performing generative LLM (Micro-F1: 0.893, Recall: 0.949), coming within 6.7 percentage points of the best supervised NER-RE solution. It was closely followed by GPT-OSS-120B (0.883 F1) and Llama-3.1-70B (0.872 F1). Notably, all five generative LLMs achieved a Global Recall exceeding 0.92, demonstrating that generative models are highly effective at retrieving clinical entities.

**Table 1**. The overall performance of the traditional NER-RE solution versus the five zero-shot open-source LLMs.

| MODEL | F1 (Micro) | PRECISION | RECALL | TP | FP | FN |
|---|---|---|---|---|---|---|
| NER_RE (Baseline) | 0.960 | 0.944 | 0.977 | 637 | 38 | 15 |
| GEMMA-3-27B | 0.831 | 0.748 | 0.936 | 610 | 206 | 42 |
| GPT-OSS-120B | 0.883 | 0.820 | **0.957** | 624 | 137 | 28 |
| GPT-OSS-20B | **0.893** | **0.843** | 0.949 | 619 | 115 | 33 |
| GRANITE-3.3-8B | 0.834 | 0.760 | 0.924 | 597 | 189 | 49 |
| LLAMA-3.3-70B | 0.872 | 0.804 | 0.954 | 622 | 152 | 30 |

**Table 2**. Field-level extraction performance (F1-score) across supervised and zero-shot architectures.

| Field | GEMMA-3-27B | GPT-OSS-120B | GPT-OSS-20B | GRANITE-3.3-8B | LLAMA-3.3-70B | NER_RE BASELINE |
|---|---|---|---|---|---|---|
| Focality | 0.763 | 0.847 | 0.857 | 0.773 | 0.916 | **0.962** |
| Histologic Grade | 0.941 | 0.980 | 0.971 | 0.970 | 0.959 | **0.980** |
| Histologic Type | 0.895 | 0.982 | 0.972 | 0.925 | **0.991** | **0.991** |
| LN Examined Count | 0.843 | 0.878 | 0.911 | 0.561 | 0.867 | **0.986** |
| LN Involved Count | 0.772 | 0.772 | 0.737 | 0.667 | 0.772 | **0.852** |
| Laterality | 0.838 | 0.891 | 0.907 | 0.869 | 0.889 | **0.915** |
| Margin Distance | 0.848 | 0.891 | 0.913 | 0.825 | 0.741 | **0.933** |
| Margin Status | 0.929 | 0.938 | 0.947 | 0.932 | 0.952 | **0.980** |
| Pathologic Stage | 0.952 | 0.964 | 0.970 | **0.994** | 0.915 | 0.970 |
| Tumor Site | 0.764 | 0.919 | 0.899 | 0.712 | 0.904 | **1.000** |

| Tumor Size | 0.690 | 0.789 | 0.849 | 0.789 | 0.781 | **0.931** |
|---|---|---|---|---|---|---|
| Visceral Pleural Invasion | 0.333 | 0.320 | 0.308 | 0.348 | 0.320 | **1.000** |
| Lymphovascular Invasion | 0.794 | 0.742 | 0.714 | 0.857 | 0.818 | **0.967** |

At the subcategory level, the full supervised NER-RE solution achieved the best performance across most CAP-aligned data fields. However, zero-shot LLMs were consistently competitive on core tumor characterization fields. For example, LLM performance on Histologic Type ranged from 0.895–0.991, and on Histologic Grade from 0.941–0.980. The agentic workflow demonstrated remarkable performance, especially on Pathologic Stage, a field that requires the synthesis of T, N, and M components. The zero-shot Granite-3.3-8B model achieved an F1 score of 0.994, outperforming the supervised NER-RE model (0.970). Similarly, for Histologic Type, the Llama-3.1-70B model matched the baseline's performance (0.991 F1), indicating that generative LLMs possess sufficient internal medical knowledge to classify histology zero-shot. Margin Status remained high across all LLMs (F1 0.929–0.952), and Tumor Site reached F1 0.899–0.919 for GPT-OSS-20B/120B and Llama-70B. However, the generative LLMs lagged slightly on strictly numeric fields such as Tumor Size, where the baseline (0.931 F1) outperformed the best LLM (GPT-OSS-20B: 0.849 F1).

**Table 3**. Computational efficiency analysis showing average and median inference latency per report.

| Model | LLAMA-3.3-70B | GEMMA3-27B | GPT-OSS-120B | GPT-OSS-20B | GRANITE-3.3-8B |
|---|---|---|---|---|---|
| Average time (secs) | **27.47 ±9.5** | 35.67 ±23.7 | 41.15 ±16.5 | 49.52 ±80.2 | 63.46 ±150.8 |
| Median time (secs) | 25.3 | 27.28 | 35.71 | **24.65** | 30.72 |

While the supervised NER-RE solution achieved the highest aggregate scores, it incurred a substantial "Time-to-Deploy cost", requiring the manual annotation of 286 documents and approximately 22 hours of GPU training time on an NVIDIA A100 GPU. In contrast, the zero-shot LangGraph workflow required zero annotation or training time. Inference latency for the workflow averaged between 27.5 and 63.5 seconds per report, with GPT-OSS-20B having the lowest median inference time to process 53 reports at 24.65 secs. This tradeoff suggests that zero-shot architectures can offer a viable, immediate-deployment alternative for clinical registries where the resource is limited for training fully supervised LLMs.

**Discussion**

Lung pathology extraction is an important task that requires substantial human effort. This study developed a real-world corpus using UF Health pathology reports and compared zero-shot generative LLMs with traditional solutions based on fully supervised NER and RE. Our findings suggest that the performance of zero-shot LLMs is getting closer to that of fully supervised LLMs in lung pathology information extraction, though it is still not on par. Traditional IE models prioritize text-level matching, clinical registries, and downstream analytics needs, but concept-level facts are not considered. Our evaluation framework focuses on the correctness of clinical facts, rather than text matching, where generative LLMs demonstrate strong performance. The fully supervised model is based on multi-stage architecture (NER followed by RE), creating a pipeline cascade where error propagation at the span level renders subsequent extraction impossible. In contrast, generative LLMs solved the IE problem using a unified text-to-text generation approach, directly generating target information. This generative AI-based framework is easier to adapt to clinical guidelines, such as CAP guidelines. When CAP guideline updates occur, generative AI-based solutions can modify prompts to adapt quickly. On the other side, fully supervised models based on NER and RE require significant effort, including time-consuming annotations to follow the new guidelines.

While the fully supervised LLM achieved a higher aggregate Precision (0.944), error analysis reveals that this metric is heavily impacted by annotator omission in the gold standard. For example, annotators frequently skip negative findings; the fully supervised model inherits this bias. Whereas generative LLMs function under a zero-shot setting that doesn't need human instruction. On fields such as Visceral Pleural Invasion (VPI) and Lymphovascular Invasion (LVI), LLMs achieved near-perfect Recall (1.000) but lower Precision because they systematically extracted valid negative findings (e.g., "Not identified") that humans forgot to annotate. Generative LLMs, however, could still capture this information missed by human annotators, which is reflected as a "lower" precision. Future evaluation frameworks for generative AI models should consider this scenario.

The agentic AI demonstrates potential for automated quality assurance in pathology reporting. The CAP protocols mandate the reporting of specific "synoptic" elements (e.g., Histologic Type, Visceral Pleural Invasion), yet variability in narrative text often leads to ambiguity or omission. Our results indicate that the LLMs are exhaustive abstractors and often populate fields with explicit negatives (e.g., "Lymphovascular Invasion: Not identified") even when human annotators forgot to annotate. This is an advantage in using practice agent-based AI models.

This study used open-source generative LLMs, enabling privacy-preserving deployment, reproducibility, and auditable evaluation pipelines, which are key requirements for healthcare systems. Our findings indicate that mid-sized open-source generative LLMs (8B–20B parameters), with agentic workflows, have sufficient reasoning power to perform high-fidelity clinical abstraction locally, facilitating HIPAA and GDPR constraints. Based on our findings, selection of an LLM should be primarily driven by local computational infrastructure. Highly resourced environments benefit from mid-sized models (20B–70B) for optimal synthesis of complex fields, while institutions restricted to single-GPU or edge deployments can leverage lightweight models (~8B) as a viable alternative for high-recall triage. However, as a single-center exploratory study with a limited test set, these findings should be interpreted with caution. Future work will focus on validating this architecture across multi-center datasets, expanding to additional oncologic schemas, and integrating human-in-the-loop interfaces to integrate our agentic AI in routine clinical workflows.

**Conclusion**

This exploratory study indicates that agentic generative AI can achieve promising lung pathology extraction results (F1 score of 0.893 for GPT-OSS-20B) without task-specific training data, providing a foundation for future large-scale, multi-institutional validation.

**Acknowledgment**

This study was partially supported by grants from NIAID R01AI172875, the Patient-Centered Outcomes Research Institute® (PCORI®) Award ME-2023C3-35934, the PARADIGM program awarded by the Advanced Research Projects Agency for Health (ARPA-H), National Institute on Aging U24AG098157, National Institute of Allergy and Infectious Diseases, National Heart, Lung, and Blood Institute, R01HL169277, R01HL176844, National Institute on Drug Abuse, NIDA R01DA057886, R01DA063631, and the UF Clinical and Translational Science Institute. The content is solely the responsibility of the authors and does not necessarily represent the official views of the funding institutions.